\documentclass{article}

\usepackage{microtype}
\usepackage{graphicx}
\usepackage{subcaption}
\usepackage{booktabs} %

\usepackage{hyperref}

\usepackage[preprint]{icml2026}

\usepackage{amsmath}
\usepackage{amssymb}
\usepackage{mathtools}
\usepackage{amsthm}

\usepackage[table]{xcolor}
\usepackage[capitalize,noabbrev]{cleveref}

\usepackage{marvosym}

\theoremstyle{plain}

\theoremstyle{definition}

\theoremstyle{remark}

\usepackage[textsize=tiny]{todonotes}

\usepackage{multirow}
\usepackage{booktabs}
\usepackage{xcolor}

\usepackage{graphicx}

\definecolor{lightgray}{gray}{0.95}
\definecolor{deepblue}{RGB}{70,130,180}
\definecolor{deepgray}{RGB}{119,136,153}
\definecolor{RosyBrown}{RGB}{188,143,143}
\definecolor{PeachPuff3}{RGB}{205,175,149}

\usepackage{enumitem}
\usepackage{listings}
\usepackage{xcolor} %

\lstdefinestyle{prompt}{
    basicstyle=\ttfamily\fontsize{7pt}{8pt}\selectfont,
    frame=none,
    breaklines=true,
    backgroundcolor=\color{lightgray},
    breakatwhitespace=true,
    breakindent=0pt,
    escapeinside={(*@}{@*)},
    numbers=none,
    numbersep=5pt,
    xleftmargin=5pt,
    aboveskip=2pt,
    belowskip=2pt,
}
\usepackage[most]{tcolorbox} 
\tcbset{
  aibox/.style={
    top=10pt,
    colback=white,
    enhanced,
    center,
  }
}
\newtcolorbox{AIbox}[2][]{aibox, title=#2,#1}

\icmltitlerunning{Preprint}

\begin{document}

\twocolumn[
  \icmltitle{Vision-DeepResearch Benchmark: Rethinking Visual and Textual Search \\ for Multimodal Large Language Models}

  \icmlsetsymbol{equal}{*}
  \icmlsetsymbol{leader}{$\dag$}
  \icmlsetsymbol{corresponding}{\Letter}

  \begin{icmlauthorlist}
    \textbf{Yu Zeng}$^{2 *}$\,
    \textbf{Wenxuan Huang}$^{1,3 *\dag}$\textsuperscript{\Letter}\,
    \textbf{Zhen Fang}$^{2 *}$\\
    \textbf{Shuang Chen}$^{5}$\,
    \textbf{Yufan Shen}$^{6}$\,
    \textbf{Yishuo Cai}$^{7}$\,
    \textbf{Xiaoman Wang}$^{3}$\,
    \textbf{Zhenfei Yin}$^{8}$\,
    \textbf{Lin Chen}$^{2}$\\
    \textbf{Zehui Chen}$^{2}$\,
    \textbf{Shiting Huang}$^{2}$\,
    \textbf{Yiming Zhao}$^{2}$\,
    \textbf{Xu Tang}$^{4}$\,
    \textbf{Yao Hu}$^{4}$\,
    \textbf{Philip Torr}$^{8}$\,
    \textbf{Wanli Ouyang}$^{1,9}$\,
    \textbf{Shaosheng Cao}$^{4}$\textsuperscript{\Letter}\,
    \\ $^1$CUHK MMLab\quad
    $^2$University of Science and Technology of China\quad
    $^3$East China Normal University\quad
    $^4$Xiaohongshu Inc.\quad
    $^5$The University of California, Los Angeles\quad
    $^6$Zhejiang University\quad
    $^7$Peking University\quad
    $^8$University of Oxford\quad
    $^9$Shenzhen Loop Area Institute\quad
    \\{\tt\small wxhuang0616@gmail.com (Wenxuan Huang)}
    \\ *: Equal Contribution\quad
    \dag: Project Leader\quad
    \Letter: Corresponding Author\quad
  \end{icmlauthorlist}

  \icmlkeywords{Machine Learning, ICML}

  \vskip 0.3in
  \printAffiliationsAndNotice{}
]

\begin{abstract}
Multimodal Large Language Models (MLLMs) have advanced VQA and now support Vision-DeepResearch systems that use search engines for complex visual-textual fact-finding. However, evaluating these visual and textual search abilities is still difficult, and existing benchmarks have two major limitations.
First, \textbf{existing benchmarks are not visual search-centric}: 
answers that should require visual search are often leaked through cross-textual cues in the text questions or can be inferred from the prior world knowledge in current MLLMs. Second, \textbf{overly idealized evaluation scenario}:
On the image-search side, the required information can often be obtained via near-exact matching against the full image, while the text-search side is overly direct and insufficiently challenging. To address these issues, we construct the \textit{Vision-DeepResearch benchmark (\textbf{VDR-Bench})} comprising 2,000 VQA instances. 
All questions are created via a careful, multi-stage curation pipeline and rigorous expert review, designed to assess the behavior of Vision-DeepResearch systems under realistic real-world conditions.
Moreover, to address the insufficient visual retrieval capabilities of current MLLMs, we propose a simple multi-round cropped-search workflow. This strategy is shown to effectively improve model performance in realistic visual retrieval scenarios.
Overall, our results provide practical guidance for the design of future multimodal deep-research systems. The code will be released in \url{https://github.com/Osilly/Vision-DeepResearch}.
\end{abstract}

\section{Introduction}
The rapid progress of Multimodal Large Language Models (MLLMs)~\cite{chen2024we,huang2025vision,zhao2025v2p,sharegpt4v,sharegpt4video,wang2025vrag,yu2024visrag,qi2025vcr,huang2025interleaving,han2026unicorn,zeng2025agentic,zeng2025enhancing} has enabled a new class of multimodal deep-research systems~\cite{webwatcher,mmsearch-r1,deepmmsearch-r1,deepeyesv2}, which combine image understanding, web search, and multi-hop reasoning to solve complex visual–textual queries. 
These systems promise to automate deep visual fact-finding by grounding reasoning in real-world images and external knowledge sources. 
Despite this progress, evaluating the visual and textual search capabilities of multimodal deep-research systems remains challenging. 
Existing benchmarks for multimodal search and knowledge-seeking VQA~\cite{webwatcher,mmsearch,fu2025livevqa,wang2017fvqa,cheng2025simplevqa,chen2023can} suffer from two fundamental limitations, as illustrated in Fig.~\ref{fig:motivation}:

\begin{figure*}[ht]
    \centering    \includegraphics[width=1.0\textwidth]{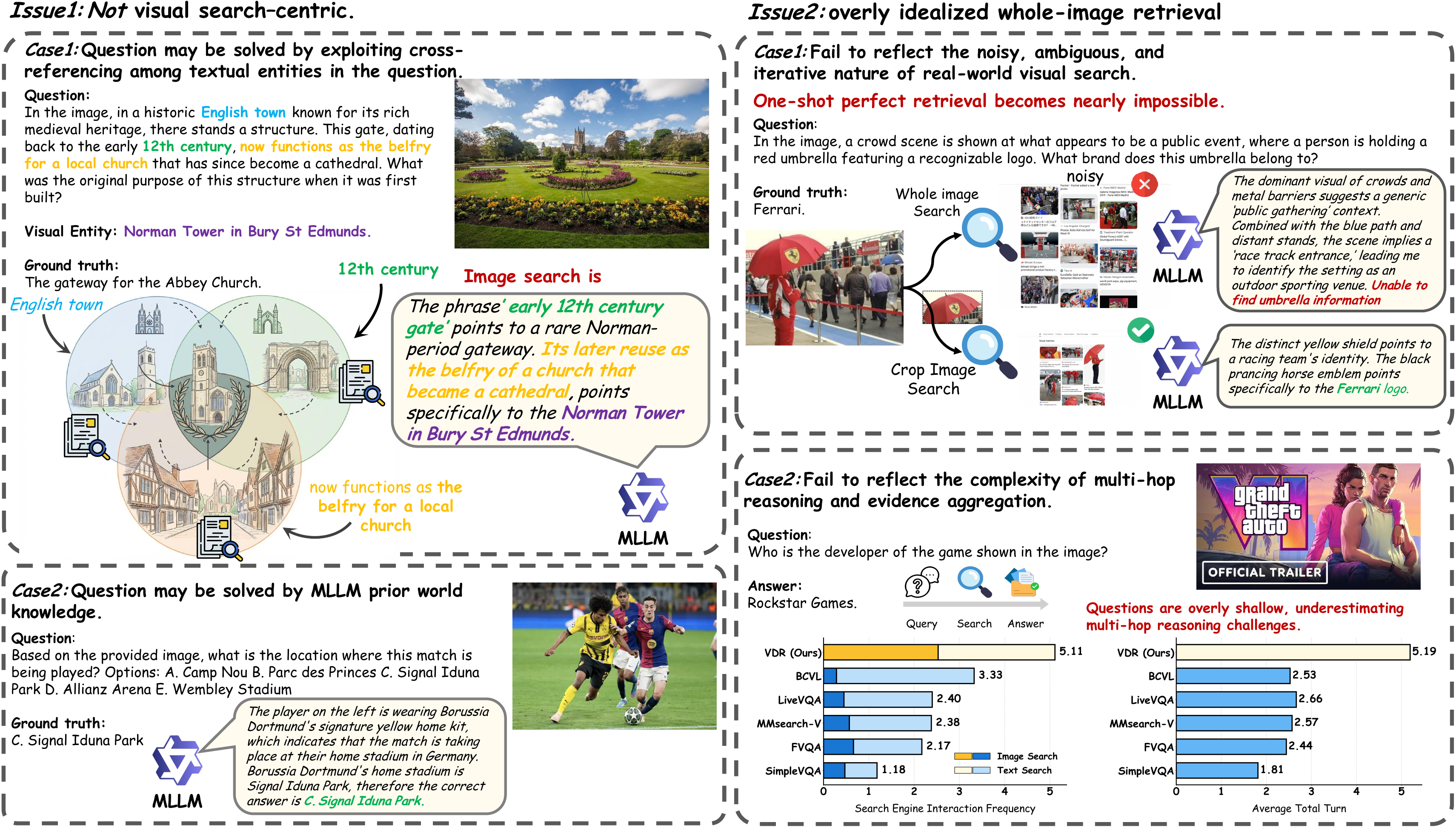}
    \caption{\textbf{Motivation}. Existing Vision-DeepResearch benchmarks often fail to measure realistic multimodal search: many questions can be solved via text-only cues or model priors without genuine visual verification, and whole-image search frequently retrieves near-duplicate images with identifying metadata (``perfect retrieval"). VDR-Bench is designed to be visual-search–centric and to reflect real-world settings that require iterative, entity-level localization (e.g., multi-round cropping), cross-modal evidence collection, and multi-hop reasoning.}
    \label{fig:motivation}
\end{figure*}

First, current benchmarks are not visual search–centric. 
Many benchmark questions can be answered without genuine visual verification, \textit{i.e.}, some answers are inferred from language priors with cross-textual cues, and even MLLM's world knowledge.
As a simple example (shown in Fig~\ref{fig:motivation}), some VQA problems in these benchmarks can be resolved by exploiting cross-referencing among textual entities in the question: taking the intersection of multiple candidate sets that satisfy different textual constraints typically yields a small set, making it easy to pin down (or even guess) the answer. 
We view this as a shortcut phenomenon.
Furthermore, as presented in Fig~\ref{fig:motivation}, some VQA instance can simply solved by MLLM prior world knowledge.
As a result, benchmark performance often reflects memorization or text-only retrieval rather than the the necessary for gathering visual information.

Second, the evaluation scenario of existing benchmarks rely on overly idealized search settings. 
On the image-search side, required information is frequently obtained through near-exact whole-image matching, where querying the original image retrieves an identical or near-duplicate source along with its title or metadata.
On the text-search side, questions are often formulated in a direct and shallow manner, underestimating the difficulty of multi-hop reasoning and evidence aggregation. 
These conditions fail to reflect the noisy, ambiguous, and iterative nature of real-world visual search. 
In realistic scenarios, visual search is inherently trial-based and iterative. 
Images often contain multiple entities, background distractors, and ambiguous visual cues, making whole-image retrieval unreliable. Effective multimodal deep-research systems should therefore localize candidate entities, issue multiple cropped image queries, refine hypotheses across rounds, and verify results through cross-modal reasoning, in order to adapt to real-world requirement.
Multi-round, multi-scale cropping is not merely an implementation detail but a necessary strategy for robust visual retrieval.

To address these challenges, we introduce \textbf{VDR-Bench} (Vision-DeepResearch Benchmark), a large-scale benchmark comprising 2,000 carefully curated VQA instances designed to evaluate realistic multimodal search behavior. 
VDR-Bench is constructed through a rigorous multi-stage pipeline that emphasizes visual-first entity discovery, human verification, and multi-hop query expansion, ensuring that solving each instance genuinely requires both visual search and textual reasoning. 
Unlike prior benchmarks that convert text-based QA into VQA through single-step image retrieval, our benchmark avoids shortcut-prone instances and minimizes perfect-retrieval bias. 
Furthermore, to improve the insufficient visual retrieval capabilities of current MLLMs, we propose a simple yet effective multi-round cropped-search workflow. 
This strategy iteratively refines visual queries by cropping regions of interest, enabling more accurate entity localization and reducing retrieval noise. 
Experimental results demonstrate that this workflow substantially improves performance on realistic visual retrieval tasks, highlighting a practical direction for building more capable multimodal deep-research systems.

Our main contributions are summarized as follows:
\begin{itemize}
    \item We identify critical limitations in existing multimodal search benchmarks, including the lack of \textbf{visual-search–centric evaluation} and \textbf{overly idealized retrieval settings}.
    \item We introduce \textbf{VDR-Bench}, a new benchmark with 2,000 instances curated through a multi-stage, human-verified pipeline that enforces realistic visual and textual search requirements.
    \item We propose a \textbf{multi-round cropped-search workflow} that significantly improves model performance in realistic visual retrieval scenarios, offering practical guidance for future multimodal deep-research systems.
\end{itemize}

\section{Related Work}
\subsection{Vision-DeepResearch Systems.}
Vision-DeepResearch agents are designed to autonomously search, read, and synthesize knowledge from the open web through multi-step reasoning and cross-modal grounding. While recent progress in text-only deep research agents~\cite{team2025tongyi,wu2025webdancer,li2025websailor,tao2025webshaper} has been substantial, the increasing visual complexity of real-world web environments has driven a paradigm shift toward multimodal Vision-DeepResearch systems. Compared with text-only counterparts, these systems face significantly greater challenges, as they must integrate visual evidence, perform reliable visual search, and conduct cross-modal verification under noisy and ambiguous conditions.

To address these challenges, several Vision-DeepResearch frameworks have been proposed. WebWatcher~\cite{webwatcher} reformulates text-based question answering into visual question answering (VQA) via reverse image search. MMSearch-R1~\cite{mmsearch-r1} enhances multimodal search strategies through Group Relative Policy Optimization (GRPO)~\cite{shao2024deepseekmath}. Building upon this line of work, DeepMMSearch-R1~\cite{deepmmsearch-r1} further improves visual grounding by employing entity-level image cropping, reducing background noise and enabling more targeted visual retrieval. Despite these advances, existing systems still struggle with the complexity and noise of realistic visual search environments.

\subsection{Evaluations of Vision-DeepResearch Systems.}
The rapid development of Vision-DeepResearch systems requires evaluation benchmarks that accurately measure visual and textual search capabilities under realistic conditions. SimpleVQA~\cite{cheng2025simplevqa} assesses multimodal factuality and hallucination robustness, while FVQA~\cite{wang2017fvqa} and InfoSeek~\cite{chen2023can} require models to ground visual evidence in external knowledge sources. LiveVQA~\cite{fu2025livevqa} evaluates reasoning over dynamic, real-time visual information, whereas MMSearch~\cite{mmsearch} and BrowseComp-VL~\cite{webwatcher} examine agentic competence in multi-step multimodal search and deep research workflows.

Although these benchmarks provide valuable progress, they exhibit two critical limitations aligned with our analysis. First, most existing benchmarks are not visual search–centric: many instances can be solved through language priors or text-only retrieval without requiring genuine visual verification, resulting in shortcut solutions that fail to evaluate true visual retrieval and grounding abilities. Second, current benchmarks rely on overly idealized search settings: on the image side, near-exact whole-image matching often enables perfect retrieval, while on the text side, queries are frequently shallow and insufficiently challenging. These design choices underestimate the noise, ambiguity, and iterative nature of real-world visual–textual search.

\section{A Quantitative Analysis of Existing Vision-DeepResearch Benchmarks}
\begin{figure*}[ht]
    \centering    \includegraphics[width=0.8\textwidth]{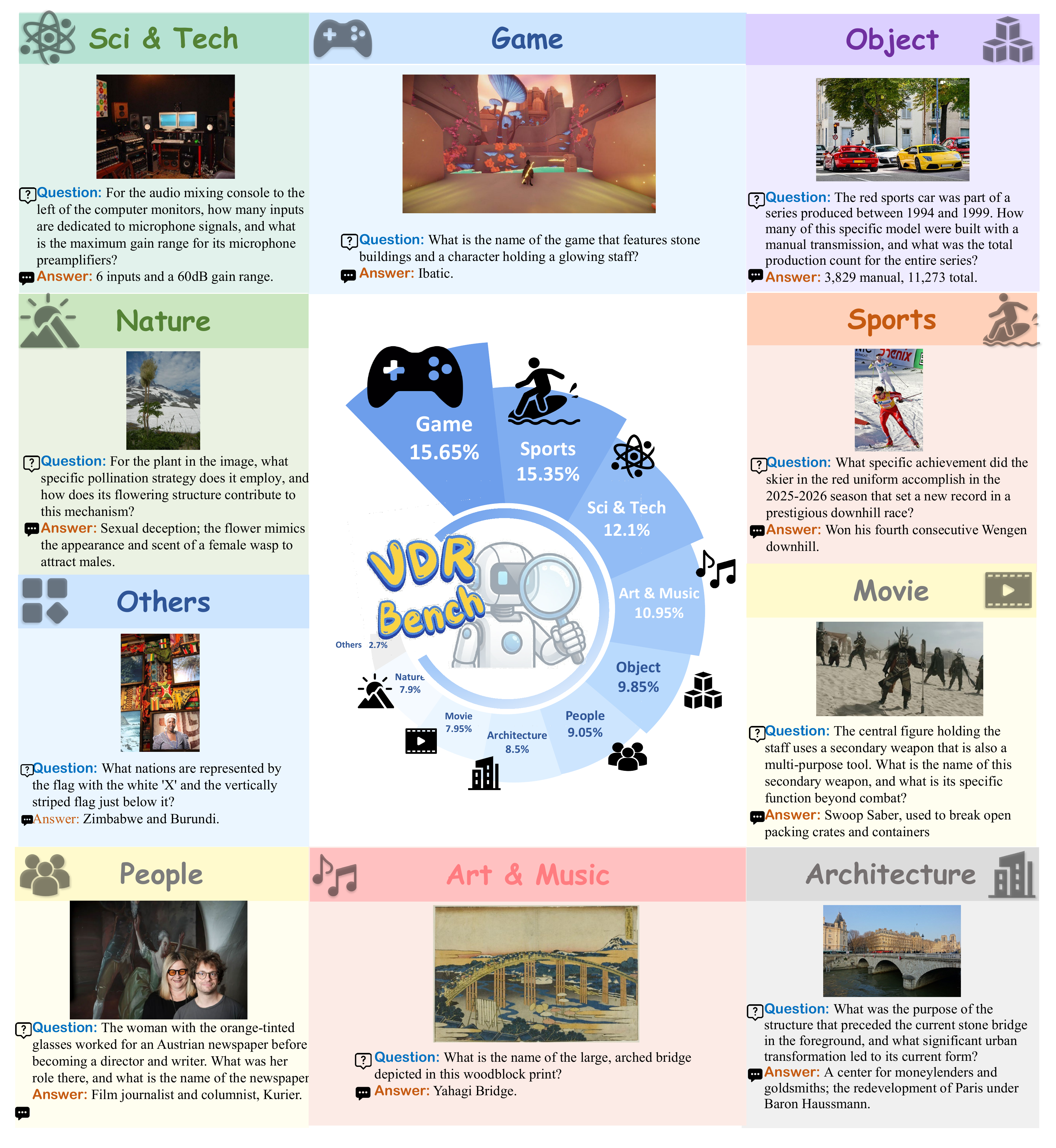}
    \caption{\textbf{Dataset composition and example instances from VDR-Bench.} The figure presents the distribution across ten visual domains, along with representative question–answer examples for each category.}
    \label{fig:mate_data}
\end{figure*}

As discussed in the Figure~\ref{fig:motivation}, existing Vision-DeepResearch benchmarks suffer from two core limitations: (1) many instances are not truly \textbf{visual-search–centric}, and (2) image retrieval is often conducted under \textbf{overly idealized evaluation scenario}. 
To quantify the severity of these issues, we conduct a controlled empirical study across representative multimodal search and knowledge-seeking VQA benchmarks, including SimpleVQA~\cite{cheng2025simplevqa}, LiveVQA~\cite{fu2025livevqa}, FVQA~\cite{wang2017fvqa}, BrowseComp-VL (BCVL)~\cite{webwatcher}, and MMSearch~\cite{mmsearch}.

We evaluate representative open and closed-source multimodal large language models, including Gemini-2.5 Pro and Qwen3-VL-30B-A3B-Instruct~\cite{2025arXiv251121631B}, as well as the text-based deep research model Tongyi-DeepResearch-30B-A3B~\cite{team2025tongyi}. 
These models are evaluated under controlled retrieval configurations to disentangle the independent contributions of whole-image search, text search, and model prior knowledge. 
We compare two experimental settings: 
(1) Given the original image and question: Direct Answer without external search, Whole-Image Search (WIS), Text Search (TS), and WIS + TS.
(2) Replacing the image with caption (produced by Qwen3-VL-235B-A3B-Instruct) alongside the question: Direct Answer and Text Search only.
This design allows us to evaluate whether benchmark instances genuinely require visual search, or can instead be solved via text-only retrieval or model priors.
The results are reported in Table~\ref{tab:performance_comparison}, where arrows ($\uparrow/\downarrow$) indicate changes relative to the No Search baseline.

\subsection{Finding 1: Existing Benchmarks Do Not Enforce Visual-Search–Centric Reasoning}
Tab.~\ref{tab:performance_comparison} reveals a clear and concerning issue: a substantial portion of instances in existing Vision-DeepResearch benchmarks can be solved without meaningful visual search. 
Across multiple datasets, enabling TS alone yields significant performance gains, and in some cases achieves results comparable to or even exceeding those obtained with Image Search. 
This indicates that many benchmark questions suffer from severe textual cue leakage, where answers can be inferred through cross-validation of textual evidence without requiring image search for visual entities.

Additionally, in several benchmarks, the Caption + Direct Answer setting attains competitive performance, suggesting that models can often bypass visual evidence entirely by relying on language priors and parametric world knowledge. 
Taken together, these findings demonstrate that many existing benchmarks fail to enforce a closed visual-evidence loop and therefore systematically overestimate models' true visual search and verification capabilities. 
Current benchmark performance often reflects proficiency in text retrieval and prior memorization, rather than genuine visual-search–centric reasoning.

\begin{table*}[ht]
\centering

\definecolor{brightgreen}{rgb}{0.0, 0.8, 0.0} 
\definecolor{brightred}{rgb}{1.0, 0.1, 0.1}   
\newcommand{\cUp}{brightgreen} 
\newcommand{\cDown}{brightred}
\newcommand{\inc}[1]{\hspace{1pt}\textcolor{\cUp}{\scriptsize\bfseries$\uparrow$#1}}
\newcommand{\dec}[1]{\hspace{1pt}\textcolor{\cDown}{\scriptsize\bfseries$\downarrow$#1}}

\caption{Performance comparison across visual search benchmarks. WIS: Whole Image Search, TS: Text Search. Arrows (\textcolor{\cUp}{$\uparrow$}/\textcolor{\cDown}{$\downarrow$}) denote performance changes relative to the corresponding \textbf{No Search} baseline. BCVL scores represent the average of Level 1 and Level 2.}

\label{tab:performance_comparison}
\setlength{\tabcolsep}{6.6pt} %
\renewcommand{\arraystretch}{1} 

\begin{tabular}{@{}l | l llll l | c@{}} 
\toprule[1.2pt]
\textbf{Type} & \textbf{Setting} & \textbf{SimpleVQA} & \textbf{LiveVQA} & \textbf{FVQA} & \textbf{BCVL} & \textbf{MMSearch}   & \textbf{VDR-Bench} \\ 
\midrule

\multicolumn{8}{l}{\cellcolor{gray!10}\textbf{Gemini 2.5 Pro}} \\ \midrule
\multirow{4}{*}{\rotatebox[origin=c]{90}{\textit{Image}}} 
& Direct Answer & 69.7 & 60.3 & 60.7 & 43.1 & 39.8 & 8.2\\
& WIS        & 75.3\inc{5.6} & 64.3\inc{4.0} & 66.7\inc{6.0} & 42.6\dec{0.5} & 50.3\inc{10.5}  & 9.8 \\
& TS         & 67.3\dec{2.4} & 72.7\inc{12.4} & 69.3\inc{8.6} & 49.9\inc{6.8} & 57.3\inc{17.5}& 5.4\\
& WIS+TS     & 74.7\inc{5.0} & 76.3\inc{16.0} & 75.7\inc{15.0} & 52.9\inc{9.8} & 69.6\inc{29.8} & 10.6\\
\cmidrule(lr){1-8}
\multirow{2}{*}{\rotatebox[origin=c]{90}{\textit{Capt.}}} 
& Direct Answer & 54.0 & 46.3 & 37.3 & 38.1 & 28.1 &4.6 \\
& TS         & 54.7\inc{0.7} & 56.0\inc{9.7} & 43.0\inc{5.7} & 43.9\inc{5.8} & 33.3\inc{5.2} & 5.2\\

\midrule

\multicolumn{8}{l}{\cellcolor{gray!10}\textbf{Qwen3-VL-30B-A3B-Instruct}} \\ \midrule
\multirow{4}{*}{\rotatebox[origin=c]{90}{\textit{Image}}} 
& Direct Answer & 56.3 & 42.7 & 34.7 & 29.6 & 18.7 &  3.8 \\
& WIS        & 67.7\inc{11.4} & 48.3\inc{5.6} & 53.7\inc{19.0} & 35.1\inc{5.5} & 36.3\inc{17.6} &  5.2\\
& TS         & 48.3\dec{8.0} & 53.3\inc{10.6} & 45.0\inc{10.3} & 37.5\inc{7.9} & 40.9\inc{22.2} &5.8 \\
& WIS+TS     & 70.0\inc{13.7} & 62.7\inc{20.0} & 73.7\inc{39.0} & 46.6\inc{17.0} & 66.7\inc{48.0} & 6.0\\
\cmidrule(lr){1-8}
\multirow{2}{*}{\rotatebox[origin=c]{90}{\textit{Capt.}}} 
& Direct Answer & 51.7 & 42.0 & 31.0 & 31.3 & 15.8  & 4.6\\
& TS         & 49.0\dec{2.7} & 46.7\inc{4.7} & 43.0\inc{12.0} & 37.9\inc{6.6} & 38.0\inc{22.2} &  7.4 \\

\midrule

\multicolumn{8}{l}{\cellcolor{gray!10}\textbf{Tongyi-DeepResearch-30B-A3B}} \\ \midrule
\multirow{2}{*}{\rotatebox[origin=c]{90}{\textit{Capt.}}} 
& Direct Answer & 49.7 & 44.7 & 31.7 & 29.4 & 14.6 &4.0\\
& TS         & 46.3\dec{3.4} & 60.7\inc{16.0} & 45.7\inc{14.0} & 44.4\inc{15.0} & 45.0\inc{30.4}  & 7.5\\
\bottomrule[1.2pt]
\end{tabular}
\end{table*}

\subsection{Finding 2: Overly Idealized Retrieval Evaluation Setting}

Our analysis further uncovers a second limitation in current benchmarks: search is often conducted under unrealistically idealized conditions to evaluate.

As shown in Tab.~\ref{tab:performance_comparison}, enabling single-shot Whole-Image Search alone frequently yields substantial performance gains relative to the Direct Answer baseline.
This behavior reveals a strong perfect-match bias, where querying the full image retrieves a near-duplicate copy accompanied by identifying titles or metadata, effectively reducing the task to a one-shot lookup. 
Under such conditions, benchmark difficulty is dominated by whether the search engine returns an exact or near-exact match, rather than evaluating an agent's ability to localize entities, refine visual queries, or verify evidence across modalities.
It is important to note that this setting departs sharply from real-world visual search, where relevant evidence rarely appears as exact duplicates and is often distributed across diverse viewpoints, crops, resolutions, and contextual scenes.
Our results indicate that many existing benchmarks contain a substantial proportion of images and metadata that can be located nearly exactly on the web, thereby overlooking the iterative localization, visual search, and entity-level cross-modal verification required in real-world scenarios.

Furthermore, as shown in Fig~\ref{fig:motivation}, most existing benchmarks in text-based multi-hop part remains insufficiently challenging, \textit{i.e.}, their text-side deep-research setups are often overly simplistic, failing to effectively stress-test the increasingly strong text deep-research capabilities of modern MLLMs.

\subsection{Why VDR-Bench?}
Our findings reveal a fundamental misalignment between existing benchmark protocols and the capabilities required for realistic multimodal deep-research systems.
Many current benchmarks allow models to succeed through text-only shortcuts, language priors, or one-shot near-duplicate retrieval, rather than necessitating iterative visual search, fine-grained localization, and cross-modal evidence aggregation.

These empirical gaps directly motivate the design of VDR-Bench. 
By emphasizing visual-search–centric reasoning with the tailored designed challenging textual search setting, mitigating perfect-retrieval bias, and evaluating agents under realistic multimodal search conditions, VDR-Bench offers a more faithful and rigorous testbed for assessing and advancing multimodal deep-research systems.

\section{VDR-Bench}
\begin{figure*}[ht]
    \centering    \includegraphics[width=1.0\textwidth]{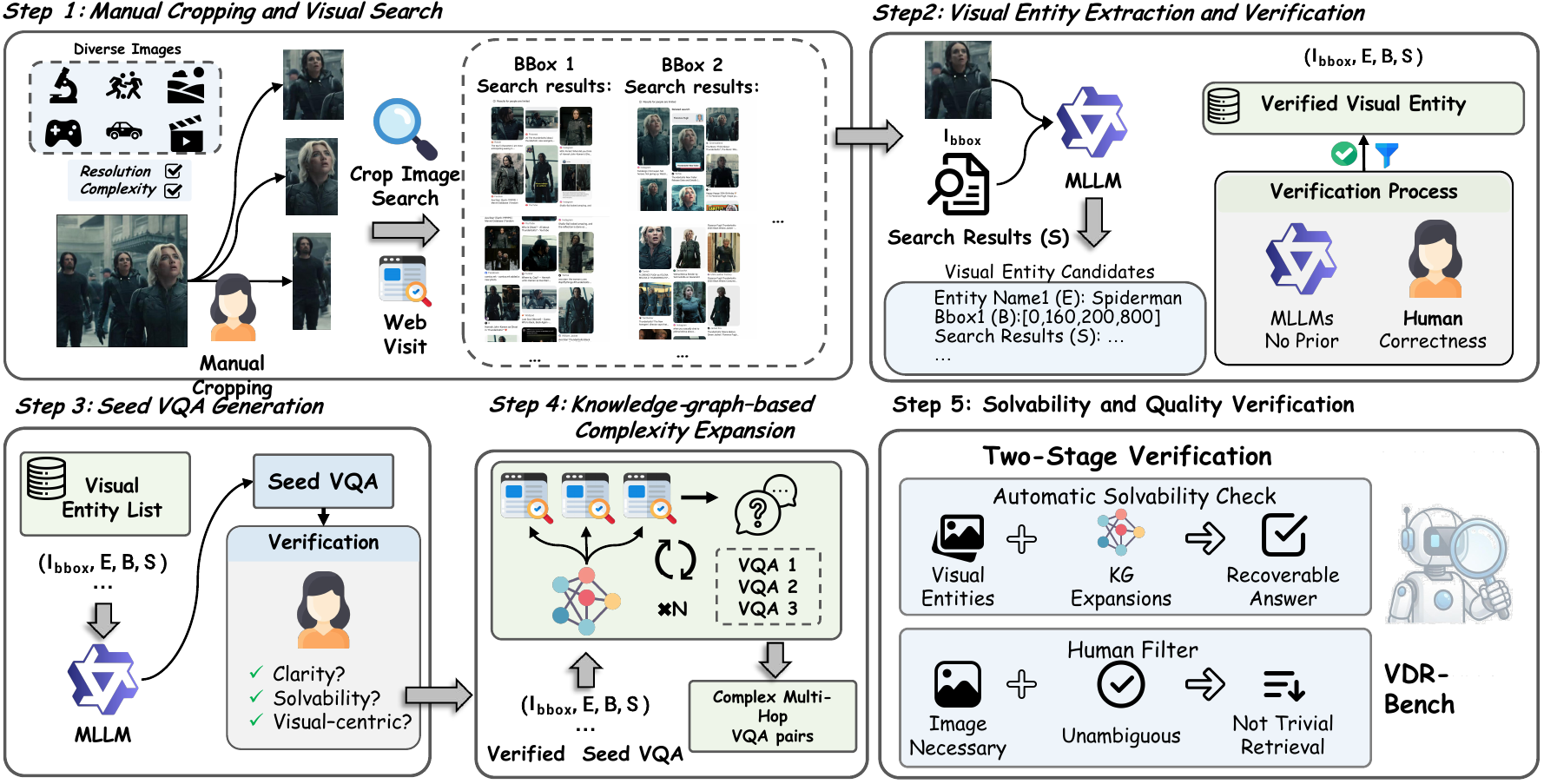}
    \caption{\textbf{VDR-Bench is constructed via a multi-stage, vision-centric workflow:} (1) annotators manually crop salient regions and perform web-scale visual search; (2) candidate entities are extracted from retrieved results and verified through an MLLM-assisted and human checking process; (3) verified visual entities are used to generate seed VQA pairs; (4) question difficulty is expanded via knowledge-graph–based multi-hop reasoning; and (5) automatic solvability checks and human quality filtering ensure that each instance requires visual evidence, remains unambiguous, and avoids trivial or near-duplicate retrieval.}
    \label{fig:data_pipeline}
\end{figure*}

In Sec.~\ref{sec:data}, we present a detailed description of the data curation process for VDR-Bench, including a multi-stage filtering pipeline and rigorous expert review.
In Sec.~\ref{sec:metrics}, we introduce two evaluation metrics specifically designed for this benchmark to assess the accuracy of deep research and entity-level retrieval performance.
\subsection{Data Curation Process}
\label{sec:data}

Our benchmark data is constructed through a strict vision-centric search pipeline that begins with raw images and culminates in multi-hop Visual Question Answering (VQA) instances.
Throughout the process, we explicitly annotate retrieved entities, thereby providing each sample with grounded visual and textual evidence.

\textbf{Step 0: Multi-Domain Image Pre-Filtering.}
We first collect images from multiple datasets spanning diverse domains and remove samples with insufficient resolution.
We then employ Qwen3-VL-235B-A22B-Instruct as an image filtering model to further select high-quality images that contain multiple entities and reflect realistic, visually rich scenes.

\textbf{Step 1: Manual Cropping and Visual Search.}
Given an original image, annotators are instructed to extract salient local regions (\textit{e.g.}, objects, logos, landmarks, or individuals) rather than relying on the full image. 
Each cropped region is subsequently used as a query to a web-scale image search engine, producing candidate visual search results.

\textbf{Step 2: Visual Entity Extraction and Verification.}
For the retrieval results associated with each crop, we extract candidate entity names (\textit{e.g.}, persons, products, locations, or organizations) from webpage titles and captions. 
Specifically, we first prompt Qwen3-VL-235B-A22B-Instruct to filter and verify the consistency between the query crop and search results, followed by human validation to ensure that entity names cannot be trivially obtained through full-image search alone.

\textbf{Step 3: Seed VQA Generation from Visual Entities.}
For each verified visual entity, we use Gemini-2.5-Pro to synthesize seed VQA pairs that require explicit recognition and grounding of the visual entity, such as identifying landmarks, brands, or object categories depicted in the image.
These VQA pairs are then manually reviewed to ensure clarity, answer uniqueness, and strict dependence on both the visual entity and its retrieved context.

\textbf{Step 4: Knowledge-Graph–Based Complexity Expansion.}
To move beyond single-hop recognition, we link each visual entity to an external knowledge graph and perform random walks to retrieve related textual entities (\textit{e.g.}, founders, cities, years, or affiliated organizations). 
We then construct multi-hop reasoning questions whose inference chains originate from the visual entity and traverse one or more knowledge-graph nodes. 
For example, questions may ask for the headquarters city of a company whose logo appears in the image, or the birth year of the architect who designed a depicted building.

\textbf{Step 5: Solvability and Quality Verification.}
All synthesized questions undergo a two-stage solvability check.
First, we automatically verify that answers can be recovered by combining the recorded visual search trajectory (cropped regions and retrieved entities) with knowledge-graph expansion, without relying on any hidden information.
Second, human annotators further filter questions to ensure the absence of text-only or prior-based shortcuts, and to confirm that reasoning paths are explicit, valid, and unambiguous.

\subsection{Benchmark Composition and Metrics}
\label{sec:metrics}
Building on the above vision-centric curation pipeline, we construct VDR-Bench, comprising $2,000$ multi-hop VQA instances spanning $10$ diverse visual domains. 
The benchmark covers a broad spectrum of visual complexity, entity density, and reasoning depth. Dataset composition and domain distribution are shown in Fig.~\ref{fig:motivation}.

We use two core metrics: answer accuracy and entity recall, to comprehensively evaluate the correctness of the answer and the search engine hit rate at the entity level.

\textbf{Answer Accuracy.}
We extract the agent's final answer from the last reasoning step and evaluate its correctness using Qwen3-VL-30B-A3B-Instruct as the judge model. 
The evaluation follows the judge prompt template from the Tongyi DeepResearch report~\cite{team2025tongyi}. 
Details are provided in the Appendix.

\textbf{Entity Recall.}
In addition to accuracy, we propose \textbf{Entity Recall (ER)} to evaluate whether a multimodal deep-research system successfully discovers the task-relevant entities and then probes deeper.
ER measures the alignment between the searched entities and a predefined \textit{gold entity sequence}.

For each task $i$, let $\mathcal{E}_i = \{e_1, \dots, e_{N_i}\}$ denote the gold entity sequence, and let $\mathcal{S}_i = \bigcup_{t=1}^{T} s_t$ denote the set of entities extracted from the agent's search trajectory across $T$ time steps. 
Specifically, ER is evaluated under an LLM-as-a-judge paradigm, which assesses the semantic equivalence between searched entities and gold annotations. 
Formally, the success of the $i$-th trajectory is defined as:
\begin{equation}
    S_i = \mathbb{1} \left[ \text{LLM}_{\text{eval}} \left( \mathcal{S}_i, \mathcal{E}_i \right) = 1 \right],
\end{equation}
where the indicator function $\mathbb{1}[\cdot]$ yields 1 if the LLM judge confirms that the set of retrieved entities $\mathcal{S}_i$ sufficiently covers the gold requirements $\mathcal{E}_i$ through semantic reasoning, and 0 otherwise. 
Unlike string matching, this approach recognizes semantic synonyms and respects the fact that an agent might follow several equally valid paths to reach the target entities.
\begin{table*}[ht]
\centering
\caption{
Performance Comparison of Models Across Different Categories (Accuracy).
Direct Answer means model directly answer the VQA without search tools, and CIS+TS means adopting both cropped-image search and text search.
MVC means the Multi-turn Visual Forcing strategy.
}
\label{tab:accuracy-results}
\footnotesize
\setlength{\tabcolsep}{3.5pt} %
\resizebox{0.9\linewidth}{!}{
\begin{tabular}{@{}lccccccccccccl@{}}
\toprule[1.3pt]
Model / Setting & People & Object & Arch. & Nature & Sci\&Tech & Art\&Music & Sports & Movie & Game & Other & Overall \\
\midrule
\midrule
\textbf{Gemini 2.5 Pro} & & & & & & & & & & & \\
Direct Answer &6.4 & 9.8 & 9.8 & 8.2 & 12.0 & 11.8 & 4.2 & 2.0 & 7.7 & 9.6 & 8.2 \\
CIS+TS &14.9 & 15.7 & 27.5 & 12.2 & 24.0 & 17.6 & 12.5 & 10.2 & 1.9 & 25.0 & 16.2 \\
\midrule
CIS+TS+MVF & 38.3 & 23.5 & 33.3 & 24.5 & 22.0 & 39.2 & 25.0 & 24.5 & 21.2 & 48.1 & 30.0 \\
\midrule
\midrule
\textbf{GPT-5} & & & & & & & & & & & \\
Direct Answer &4.4 & 9.8 & 11.7 & 12.3 & 10.0 & 7.8 & 8.4 & 8.2 & 3.8 & 13.5 & 9.5 \\

CIS+TS &20.8 & 17.6 & 14.0 & 16.7 & 24.5 & 21.2 & 12.5 & 19.3 & 20.8 & 25.0 & 19.2 \\
\midrule
CIS+TS+MVF &23.4 & 25.5 & 23.5 & 20.4 & 18.0 & 27.5 & 22.9 & 30.6 & 30.8 & 42.3 & 26.6 \\
\midrule
\midrule
\textbf{Claude-4-Sonnet} & & & & & & & & & & & \\
Direct Answer &2.1 & 3.9 & 7.8 & 6.2 & 10.0 & 7.8 & 2.2 & 0.0 & 3.8 & 11.5 & 5.6 \\CIS+TS & 14.9 & 9.8 & 19.6 & 16.3 & 18.0 & 11.8 & 10.4 & 4.1 & 3.8 & 23.1 & 13.2\\
\midrule
CIS+TS+MVF & 12.5 & 17.6 & 24.0 & 35.4 & 15.1 & 26.9 & 16.7 & 12.3 & 23.1 & 24.4 & 20.6 \\ \midrule
\midrule
\textbf{Qwen3-VL-30B-A3B-Instruct} & & & & & & & & & & & \\
Direct Answer &0.0 & 3.9 & 3.9 & 6.1 & 2.0 & 7.8 & 2.1 & 4.1 & 0.0 & 7.7 & 3.8 \\
CIS+TS & 17.0 & 19.6 & 17.6 & 16.3 & 20.0 & 5.9 & 14.6 & 10.2 & 5.8 & 44.2 & 17.2 \\
\midrule
CIS+TS+MVF & 25.5 & 21.6 & 23.5 & 18.4 & 8.0 & 23.5 & 16.7 & 18.4 & 28.8 & 26.9 & 21.2 \\
\midrule
\midrule
\textbf{Qwen3-VL-235B-A22B-Instruct} & & & & & & & & & & & \\
Direct Answer &6.2 & 3.9 & 10.0 & 22.9 & 7.5 & 13.5 & 6.2 & 3.5 & 7.5 & 7.5 & 8.8 \\
CIS+TS & 25.2 & 19.5 & 24.0 & 21.1 & 18.5 & 17.1 & 10.7 & 29.1 & 16.6 & 31.5 & 21.2 \\
\midrule
CIS+TS+MVF&25.0 & 23.5 & 30.0 & 31.2 & 30.2 & 28.8 & 20.8 & 22.8 & 30.2 & 32.5 & 27.4 \\ \bottomrule[1.2pt]
\end{tabular}
}
\end{table*}

\section{Experiments}
\subsection{Experimental Setups}
We evaluate state-of-the-art Vision–Language Models (VLMs), including Gemini 2.5 Pro~\cite{comanici2025gemini}, GPT-5~\cite{singh2025openai}, Claude-4-Sonnet~\cite{anthropic2026claude4}, Qwen3-VL-30B-A3B-Instruct~\cite{bai2025qwen2}, Qwen3-VL-235B-A22B-Instruct~\cite{bai2025qwen2},  Vision-DeepResearch~\cite{vision-deepresearch} on VDR-Bench. Performance is reported as answer accuracy across 10 visual domains with overall results summarized in Table~\ref{tab:accuracy-results}.

To isolate the impact of search and visual grounding, we benchmark each model under three controlled inference settings:

\textbf{(1) Direct Answer.} Models answer questions using only the input image and question, without external search.

\textbf{(2) CIS+TS.} Models are equipped with a vision-centric cropped-image search (CIS) module and a text search (TS) interface, allowing multi-step retrieval over both visual and textual sources.

\textbf{(3) CIS+TS+MVF.} Building on CIS+TS, we further enable Multi-turn Visual Forcing (MVF), which encourages iterative region cropping, refined visual querying, and explicit cross-modal evidence verification.

All models operate under a fixed search budget and identical interaction constraints to ensure a fair comparison. This design enables us to quantify the contribution of vision-centric search, textual retrieval, and multi-turn visual grounding to overall reasoning performance.

\subsection{Results Analysis.}
The main results are shown in Tab.~\ref{tab:accuracy-results}. We first observe that all models achieve relatively low scores when answering directly, which clearly indicates that VDR-Bench requires models to actively perform search to obtain the answers, rather than relying solely on prior knowledge.
When equipped with search tools(CIS+TS), Qwen3-VL-235B-A22B-Instruct achieves the highest score of 21.2, even outperforming all closed-source models. We refer to this phenomenon as \textbf{lazy search}: models with strong prior knowledge tend to rely on textual reasoning or avoid search tools altogether when facing visual deep-research tasks. As a result, their powerful priors do not effectively translate into improved visual search capability.
In contrast, open-source models with weaker prior knowledge demonstrate surprisingly strong search capabilities. Moreover, this ability improves steadily as model scale increases. These findings suggest that enhancing a vision deep-research agent cannot rely solely on improving the base model through pretraining; instead, it is crucial to encourage extensive and effective use of search tools when handling such tasks.

\begin{figure}[]
    \centering    \includegraphics[width=1.0\columnwidth]{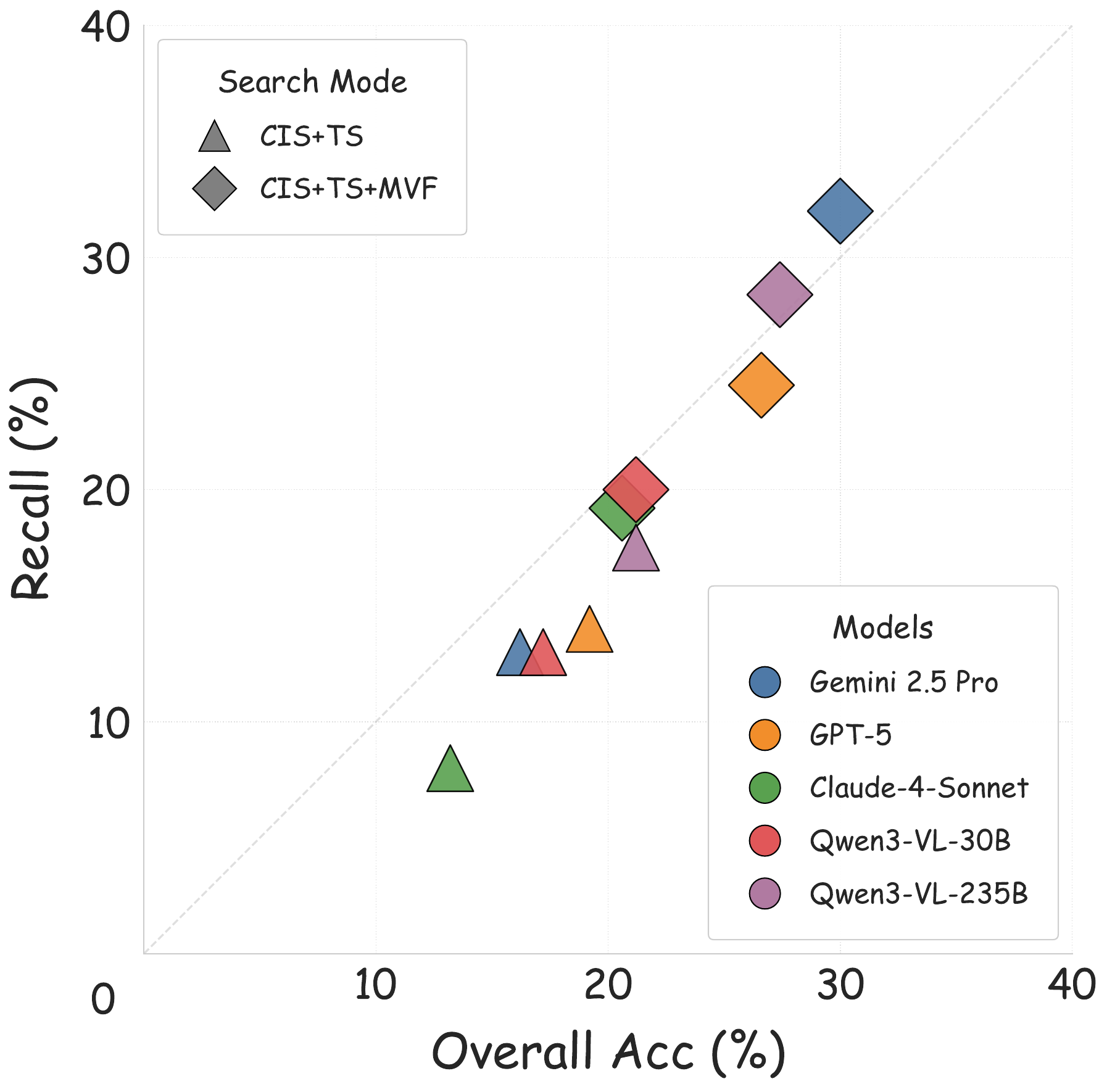}
    \caption{Relationship between overall answer accuracy and entity-level recall on VDR-Bench. Points correspond to different models evaluated under two retrieval strategies (CIS+TS and CIS+TS+MVF), with colors indicating model families and marker shapes indicating the search mode. The plot shows a strong positive association between correctly answering questions and successfully retrieving task-relevant entities, and further indicates that MVF tends to improve both metrics across models.}
    \label{fig:corr}
\end{figure}

\subsection{The Correct Paradigm and Challenges.}
The existence of the lazy search phenomenon raises an important question: why do models’ strong prior knowledge and general capabilities fail to transfer effectively to vision deep-research tasks?

To address this issue, we propose Multi-turn Visual Forcing (MVC), a zero-shot method designed to enhance models’ performance on vision deep-research tasks. MVC first guides the model to conduct fine-grained, multi-scale visual retrieval over the input image, and then encourages deeper reasoning grounded in the retrieved visual evidence, thereby better leveraging the model’s prior knowledge. Through iterative interaction with the search engine, the model becomes more likely to acquire relevant world knowledge about key entities.

Tab~\ref{tab:accuracy-results} show that MVC effectively improves the vision deep-research capabilities of different models.
Among them, Gemini achieves the highest score($16.2 \to 30.0$).

\section{Conclusion}
We identify two critical flaws in existing Vision-DeepResearch benchmarks: reliance on text-only or prior-based shortcuts, and overly idealized one-shot image retrieval that fails to reflect realistic visual search. These issues hinder faithful evaluation of visual grounding and cross-modal verification. To overcome these limitations, we propose VDR-Bench, a vision-first benchmark built with multi-scale cropping, entity-level verification, and knowledge-graph–based multi-hop reasoning. VDR-Bench enforces visual necessity and mitigates shortcut exploitation, offering a more realistic testbed for Vision-DeepResearch systems. Experiments show that strong performance depends on iterative visual search and cross-modal evidence aggregation, and that a simple multi-round cropped search strategy already yields meaningful gains. Our results provide practical guidance for building more robust multimodal deep research agents.

\newpage
\bibliography{example_paper}

\begin{thebibliography}{32}
\providecommand{\natexlab}[1]{#1}
\providecommand{\url}[1]{\texttt{#1}}
\expandafter\ifx\csname urlstyle\endcsname\relax
  \providecommand{\doi}[1]{doi: #1}\else
  \providecommand{\doi}{doi: \begingroup \urlstyle{rm}\Url}\fi

\bibitem[{Anthropic}(2025)]{anthropic2026claude4}
{Anthropic}.
\newblock Introducing claude 4.
\newblock \url{https://www.anthropic.com/news/claude-4}, 2025.

\bibitem[{Bai} et~al.(2025){Bai}, {Cai}, {Chen}, {Chen}, {Chen}, {Cheng}, {Deng}, {Ding}, {Gao}, {Ge}, {Ge}, {Guo}, {Huang}, {Huang}, {Huang}, {Hui}, {Jiang}, {Li}, {Li}, {Li}, {Li}, {Lin}, {Lin}, {Liu}, {Liu}, {Liu}, {Liu}, {Liu}, {Liu}, {Lu}, {Luo}, {Lv}, {Men}, {Meng}, {Ren}, {Ren}, {Song}, {Sun}, {Tang}, {Tu}, {Wan}, {Wang}, {Wang}, {Wang}, {Wang}, {Xie}, {Xu}, {Xu}, {Xu}, {Yang}, {Yang}, {Yang}, {Yang}, {Yu}, {Zhang}, {Zhang}, {Zhang}, {Zheng}, {Zhong}, {Zhou}, {Zhou}, {Zhou}, {Zhu}, and {Zhu}]{2025arXiv251121631B}
{Bai}, S., {Cai}, Y., {Chen}, R., {Chen}, K., {Chen}, X., {Cheng}, Z., {Deng}, L., {Ding}, W., {Gao}, C., {Ge}, C., {Ge}, W., {Guo}, Z., {Huang}, Q., {Huang}, J., {Huang}, F., {Hui}, B., {Jiang}, S., {Li}, Z., {Li}, M., {Li}, M., {Li}, K., {Lin}, Z., {Lin}, J., {Liu}, X., {Liu}, J., {Liu}, C., {Liu}, Y., {Liu}, D., {Liu}, S., {Lu}, D., {Luo}, R., {Lv}, C., {Men}, R., {Meng}, L., {Ren}, X., {Ren}, X., {Song}, S., {Sun}, Y., {Tang}, J., {Tu}, J., {Wan}, J., {Wang}, P., {Wang}, P., {Wang}, Q., {Wang}, Y., {Xie}, T., {Xu}, Y., {Xu}, H., {Xu}, J., {Yang}, Z., {Yang}, M., {Yang}, J., {Yang}, A., {Yu}, B., {Zhang}, F., {Zhang}, H., {Zhang}, X., {Zheng}, B., {Zhong}, H., {Zhou}, J., {Zhou}, F., {Zhou}, J., {Zhu}, Y., and {Zhu}, K.
\newblock {Qwen3-VL Technical Report}.
\newblock \emph{arXiv e-prints}, art. arXiv:2511.21631, November 2025.
\newblock \doi{10.48550/arXiv.2511.21631}.

\bibitem[Bai et~al.(2025)Bai, Chen, Liu, Wang, Ge, Song, Dang, Wang, Wang, Tang, et~al.]{bai2025qwen2}
Bai, S., Chen, K., Liu, X., Wang, J., Ge, W., Song, S., Dang, K., Wang, P., Wang, S., Tang, J., et~al.
\newblock Qwen2. 5-vl technical report.
\newblock \emph{arXiv preprint arXiv:2502.13923}, 2025.

\bibitem[Chen et~al.(2024{\natexlab{a}})Chen, Li, Dong, Zhang, He, Wang, Zhao, and Lin]{sharegpt4v}
Chen, L., Li, J., Dong, X., Zhang, P., He, C., Wang, J., Zhao, F., and Lin, D.
\newblock Sharegpt4v: Improving large multi-modal models with better captions.
\newblock In \emph{European Conference on Computer Vision}, pp.\  370--387. Springer, 2024{\natexlab{a}}.

\bibitem[Chen et~al.(2024{\natexlab{b}})Chen, Li, Dong, Zhang, Zang, Chen, Duan, Wang, Qiao, Lin, et~al.]{chen2024we}
Chen, L., Li, J., Dong, X., Zhang, P., Zang, Y., Chen, Z., Duan, H., Wang, J., Qiao, Y., Lin, D., et~al.
\newblock Are we on the right way for evaluating large vision-language models?
\newblock \emph{Advances in Neural Information Processing Systems}, 37:\penalty0 27056--27087, 2024{\natexlab{b}}.

\bibitem[Chen et~al.(2024{\natexlab{c}})Chen, Wei, Li, Dong, Zhang, Zang, Chen, Duan, Tang, Yuan, et~al.]{sharegpt4video}
Chen, L., Wei, X., Li, J., Dong, X., Zhang, P., Zang, Y., Chen, Z., Duan, H., Tang, Z., Yuan, L., et~al.
\newblock Sharegpt4video: Improving video understanding and generation with better captions.
\newblock \emph{Advances in Neural Information Processing Systems}, 37:\penalty0 19472--19495, 2024{\natexlab{c}}.

\bibitem[Chen et~al.(2023)Chen, Hu, Luan, Sun, Changpinyo, Ritter, and Chang]{chen2023can}
Chen, Y., Hu, H., Luan, Y., Sun, H., Changpinyo, S., Ritter, A., and Chang, M.-W.
\newblock Can pre-trained vision and language models answer visual information-seeking questions?
\newblock \emph{arXiv preprint arXiv:2302.11713}, 2023.

\bibitem[Cheng et~al.(2025)Cheng, Zhang, Zhang, Yang, Guan, Wu, Li, Zhang, Liu, Mai, et~al.]{cheng2025simplevqa}
Cheng, X., Zhang, W., Zhang, S., Yang, J., Guan, X., Wu, X., Li, X., Zhang, G., Liu, J., Mai, Y., et~al.
\newblock Simplevqa: Multimodal factuality evaluation for multimodal large language models.
\newblock In \emph{Proceedings of the IEEE/CVF International Conference on Computer Vision}, pp.\  4637--4646, 2025.

\bibitem[Comanici et~al.(2025)Comanici, Bieber, Schaekermann, Pasupat, Sachdeva, Dhillon, Blistein, Ram, Zhang, Rosen, et~al.]{comanici2025gemini}
Comanici, G., Bieber, E., Schaekermann, M., Pasupat, I., Sachdeva, N., Dhillon, I., Blistein, M., Ram, O., Zhang, D., Rosen, E., et~al.
\newblock Gemini 2.5: Pushing the frontier with advanced reasoning, multimodality, long context, and next generation agentic capabilities.
\newblock \emph{arXiv preprint arXiv:2507.06261}, 2025.

\bibitem[Fu et~al.(2025)Fu, Peng, Liu, Wan, and Chen]{fu2025livevqa}
Fu, M., Peng, Y., Liu, B., Wan, Y., and Chen, D.
\newblock Livevqa: Live visual knowledge seeking.
\newblock \emph{arXiv preprint arXiv:2504.05288}, 2025.

\bibitem[Geng et~al.(2025)Geng, Xia, Zhang, Wang, Wang, Ding, Wang, Wu, Zhao, Li, et~al.]{webwatcher}
Geng, X., Xia, P., Zhang, Z., Wang, X., Wang, Q., Ding, R., Wang, C., Wu, J., Zhao, Y., Li, K., et~al.
\newblock Webwatcher: Breaking new frontier of vision-language deep research agent.
\newblock \emph{arXiv preprint arXiv:2508.05748}, 2025.

\bibitem[Han et~al.(2026)Han, Fang, Sun, Ma, Wang, Zeng, Chen, Chen, Huang, Xu, et~al.]{han2026unicorn}
Han, R., Fang, Z., Sun, X., Ma, Y., Wang, Z., Zeng, Y., Chen, Z., Chen, L., Huang, W., Xu, W.-J., et~al.
\newblock Unicorn: Towards self-improving unified multimodal models through self-generated supervision.
\newblock \emph{arXiv preprint arXiv:2601.03193}, 2026.

\bibitem[Hong et~al.(2025)Hong, Zhao, Zhu, Lu, Xu, and Yu]{deepeyesv2}
Hong, J., Zhao, C., Zhu, C., Lu, W., Xu, G., and Yu, X.
\newblock Deepeyesv2: Toward agentic multimodal model.
\newblock \emph{arXiv preprint arXiv:2511.05271}, 2025.

\bibitem[Huang et~al.(2025{\natexlab{a}})Huang, Chen, Xie, Cao, Tang, Shen, Yin, Hu, Wang, Tang, et~al.]{huang2025interleaving}
Huang, W., Chen, S., Xie, Z., Cao, S., Tang, S., Shen, Y., Yin, Q., Hu, W., Wang, X., Tang, Y., et~al.
\newblock Interleaving reasoning for better text-to-image generation.
\newblock \emph{arXiv preprint arXiv:2509.06945}, 2025{\natexlab{a}}.

\bibitem[Huang et~al.(2025{\natexlab{b}})Huang, Jia, Zhai, Cao, Ye, Zhao, Xu, Hu, and Lin]{huang2025vision}
Huang, W., Jia, B., Zhai, Z., Cao, S., Ye, Z., Zhao, F., Xu, Z., Hu, Y., and Lin, S.
\newblock Vision-r1: Incentivizing reasoning capability in multimodal large language models.
\newblock \emph{arXiv preprint arXiv:2503.06749}, 2025{\natexlab{b}}.

\bibitem[Huang et~al.(2026)Huang, Zeng, Wang, Fang, Cao, Chu, Yin, Chen, Yin, Chen, Chen, Hu, Torr, Zhao, and Ouyang]{vision-deepresearch}
Huang, W., Zeng, Y., Wang, Q., Fang, Z., Cao, S., Chu, Z., Yin, Q., Chen, S., Yin, Z., Chen, L., Chen, Z., Hu, Y., Torr, P., Zhao, F., and Ouyang, W.
\newblock Vision-deepresearch: Incentivizing deepresearch capability in multimodal large language models.
\newblock \emph{preprint}, 2026.

\bibitem[Jiang et~al.()Jiang, Zhang, Guo, Wu, Qiu, Lu, Chen, Song, Gao, Liu, et~al.]{mmsearch}
Jiang, D., Zhang, R., Guo, Z., Wu, Y., Qiu, P., Lu, P., Chen, Z., Song, G., Gao, P., Liu, Y., et~al.
\newblock Mmsearch: Unveiling the potential of large models as multi-modal search engines.
\newblock In \emph{The Thirteenth International Conference on Learning Representations}.

\bibitem[Li et~al.(2025)Li, Zhang, Yin, Zhang, Ou, Wu, Yin, Li, Tao, Wang, et~al.]{li2025websailor}
Li, K., Zhang, Z., Yin, H., Zhang, L., Ou, L., Wu, J., Yin, W., Li, B., Tao, Z., Wang, X., et~al.
\newblock Websailor: Navigating super-human reasoning for web agent.
\newblock \emph{arXiv preprint arXiv:2507.02592}, 2025.

\bibitem[Narayan et~al.(2025)Narayan, Xu, Cao, Nerella, Patel, Shiee, Grasch, Jia, Yang, and Gan]{deepmmsearch-r1}
Narayan, K., Xu, Y., Cao, T., Nerella, K., Patel, V.~M., Shiee, N., Grasch, P., Jia, C., Yang, Y., and Gan, Z.
\newblock Deepmmsearch-r1: Empowering multimodal llms in multimodal web search.
\newblock \emph{arXiv preprint arXiv:2510.12801}, 2025.

\bibitem[OpenAI.(2025)]{singh2025openai}
OpenAI.
\newblock Openai gpt-5 system card.
\newblock \emph{arXiv preprint arXiv:2601.03267}, 2025.

\bibitem[Qi et~al.(2025)Qi, Zhao, Zeng, Bao, Huang, Chen, Chen, Zhao, Qi, and Zhao]{qi2025vcr}
Qi, Y., Zhao, Y., Zeng, Y., Bao, X., Huang, W., Chen, L., Chen, Z., Zhao, J., Qi, Z., and Zhao, F.
\newblock Vcr-bench: A comprehensive evaluation framework for video chain-of-thought reasoning.
\newblock \emph{arXiv preprint arXiv:2504.07956}, 2025.

\bibitem[Shao et~al.(2024)Shao, Wang, Zhu, Xu, Song, Bi, Zhang, Zhang, Li, Wu, et~al.]{shao2024deepseekmath}
Shao, Z., Wang, P., Zhu, Q., Xu, R., Song, J., Bi, X., Zhang, H., Zhang, M., Li, Y., Wu, Y., et~al.
\newblock Deepseekmath: Pushing the limits of mathematical reasoning in open language models.
\newblock \emph{arXiv preprint arXiv:2402.03300}, 2024.

\bibitem[Tao et~al.(2025)Tao, Wu, Yin, Zhang, Li, Shen, Li, Zhang, Wang, Jiang, et~al.]{tao2025webshaper}
Tao, Z., Wu, J., Yin, W., Zhang, J., Li, B., Shen, H., Li, K., Zhang, L., Wang, X., Jiang, Y., et~al.
\newblock Webshaper: Agentically data synthesizing via information-seeking formalization.
\newblock \emph{arXiv preprint arXiv:2507.15061}, 2025.

\bibitem[Team et~al.(2025)Team, Li, Zhang, Zhang, Huang, Li, Chen, Yin, Wu, Zhou, et~al.]{team2025tongyi}
Team, T.~D., Li, B., Zhang, B., Zhang, D., Huang, F., Li, G., Chen, G., Yin, H., Wu, J., Zhou, J., et~al.
\newblock Tongyi deepresearch technical report.
\newblock \emph{arXiv preprint arXiv:2510.24701}, 2025.

\bibitem[Wang et~al.(2017)Wang, Wu, Shen, Dick, and Van Den~Hengel]{wang2017fvqa}
Wang, P., Wu, Q., Shen, C., Dick, A., and Van Den~Hengel, A.
\newblock Fvqa: Fact-based visual question answering.
\newblock \emph{IEEE transactions on pattern analysis and machine intelligence}, 40\penalty0 (10):\penalty0 2413--2427, 2017.

\bibitem[Wang et~al.(2025)Wang, Ding, Zeng, Chen, Chen, Wang, Xie, Huang, and Zhao]{wang2025vrag}
Wang, Q., Ding, R., Zeng, Y., Chen, Z., Chen, L., Wang, S., Xie, P., Huang, F., and Zhao, F.
\newblock Vrag-rl: Empower vision-perception-based rag for visually rich information understanding via iterative reasoning with reinforcement learning.
\newblock \emph{arXiv preprint arXiv:2505.22019}, 2025.

\bibitem[Wu et~al.(2025{\natexlab{a}})Wu, Deng, Li, Liu, You, Li, Ma, and Liu]{mmsearch-r1}
Wu, J., Deng, Z., Li, W., Liu, Y., You, B., Li, B., Ma, Z., and Liu, Z.
\newblock Mmsearch-r1: Incentivizing lmms to search.
\newblock \emph{arXiv preprint arXiv:2506.20670}, 2025{\natexlab{a}}.

\bibitem[Wu et~al.(2025{\natexlab{b}})Wu, Li, Fang, Yin, Zhang, Tao, Zhang, Xi, Fu, Jiang, et~al.]{wu2025webdancer}
Wu, J., Li, B., Fang, R., Yin, W., Zhang, L., Tao, Z., Zhang, D., Xi, Z., Fu, G., Jiang, Y., et~al.
\newblock Webdancer: Towards autonomous information seeking agency.
\newblock \emph{arXiv preprint arXiv:2505.22648}, 2025{\natexlab{b}}.

\bibitem[Yu et~al.(2024)Yu, Tang, Xu, Cui, Ran, Yan, Liu, Wang, Han, Liu, et~al.]{yu2024visrag}
Yu, S., Tang, C., Xu, B., Cui, J., Ran, J., Yan, Y., Liu, Z., Wang, S., Han, X., Liu, Z., et~al.
\newblock Visrag: Vision-based retrieval-augmented generation on multi-modality documents.
\newblock \emph{arXiv preprint arXiv:2410.10594}, 2024.

\bibitem[Zeng et~al.(2025{\natexlab{a}})Zeng, Huang, Huang, Bao, Qi, Zhao, Wang, Chen, Chen, Chen, et~al.]{zeng2025agentic}
Zeng, Y., Huang, W., Huang, S., Bao, X., Qi, Y., Zhao, Y., Wang, Q., Chen, L., Chen, Z., Chen, H., et~al.
\newblock Agentic jigsaw interaction learning for enhancing visual perception and reasoning in vision-language models.
\newblock \emph{arXiv preprint arXiv:2510.01304}, 2025{\natexlab{a}}.

\bibitem[Zeng et~al.(2025{\natexlab{b}})Zeng, Qi, Zhao, Bao, Chen, Chen, Huang, Zhao, and Zhao]{zeng2025enhancing}
Zeng, Y., Qi, Y., Zhao, Y., Bao, X., Chen, L., Chen, Z., Huang, S., Zhao, J., and Zhao, F.
\newblock Enhancing large vision-language models with ultra-detailed image caption generation.
\newblock In \emph{Proceedings of the 2025 Conference on Empirical Methods in Natural Language Processing}, pp.\  26703--26729, 2025{\natexlab{b}}.

\bibitem[Zhao et~al.(2025)Zhao, Zeng, Qi, Liu, Chen, Chen, Bao, Zhao, and Zhao]{zhao2025v2p}
Zhao, Y., Zeng, Y., Qi, Y., Liu, Y., Chen, L., Chen, Z., Bao, X., Zhao, J., and Zhao, F.
\newblock V2p-bench: Evaluating video-language understanding with visual prompts for better human-model interaction.
\newblock \emph{arXiv preprint arXiv:2503.17736}, 2025.

\end{thebibliography}
\bibliographystyle{icml2026}

\newpage
\appendix
\onecolumn

\section{More Results}
\begin{table*}[ht]
\centering
\caption{
Performance comparison of models across different categories (Accuracy) on testmini set.
Direct Answer means the model answers VQA directly without search tools.
CIS+TS means adopting both the cropped-image search and text search.
}
\label{tab:accuracy-results-testmini}
\footnotesize
\setlength{\tabcolsep}{4pt}
\resizebox{0.95\linewidth}{!}{
\begin{tabular}{lccccccccccc}
\toprule[1.2pt]
Model / Setting & People & Object & Arch. & Nature & Sci\&Tech & Art\&Music & Sports & Movie & Game & Other & Overall \\
\midrule

\textbf{Gemini 2.5 Pro} \\
Direct Answer &
12.5 & 2.0 & 8.0 & 8.3 & 0.0 & 15.4 & 6.3 & 5.3 & 9.4 & 15.0& 8.0
 \\
CIS+TS 
& 22.9 & 21.6 & 14.0 & 16.7 & 17.0 & 17.3 & 20.8 & 21.1 & 11.3 & 22.5 & 18.8\\

\midrule

\textbf{GPT-5} \\
Direct Answer 
& 6.3 & 9.8 & 12.0 & 12.5 & 11.3 & 7.7 & 8.3 & 8.8 & 9.4 & 12.5 & 9.5 \\
CIS+TS 
&14.6 & 27.5 & 20.0 & 29.2 & 15.1 & 26.9 & 10.4 & 14.0 & 11.3 & 32.5& 20.4\\

\midrule

\textbf{Claude-4-Sonnet} \\
Direct Answer 
& 0.0 & 0.0 & 0.0 & 0.0 & 0.0 & 0.0 & 0.0 & 1.8 & 3.8 & 12.5 & 2.0 \\
CIS+TS 
&18.8 & 11.8 & 14.0 & 18.8 & 13.2 & 13.5 & 8.3 & 5.3 & 3.8 & 22.5& 13.6\\

\midrule

\textbf{Qwen3-VL-30B-A3B-Instruct} \\
Direct Answer 
& 2.1 & 2.0 & 6.0 & 4.2 & 3.8 & 5.8 & 2.1 & 3.5 & 7.5 & 7.5 & 3.8 \\
CIS+TS 
&12.5 & 25.5 & 16.0 & 22.9 & 15.1 & 19.2 & 10.4 & 24.6 & 3.8 & 45.0 & 20.2 \\

\midrule

\textbf{Qwen3-VL-235B-A22B-Instruct} \\
Direct Answer &
6.3 & 2.0 & 8.0 & 22.9 & 1.9 & 9.6 & 8.3 & 3.5 & 13.2 & 7.5& 7.9 \\
CIS+TS 
& 25.0 & 13.7 & 22.0 & 20.8 & 17.0 & 17.3 & 8.3 & 26.3 & 15.1 & 45.0& 20.8 \\

\midrule

\textbf{Vision-DeepResearch-8B} \\
Direct Answer 
& 6.3 & 0.0 & 10.0 & 2.1 & 0.0 & 3.8 & 0.0 & 0.0 & 0.0 & 12.5 & 3.2 \\
CIS+TS 
& 24.5 & 35.8 & 21.2 & 38.4 & 18.6 & 33.1 & 27.7 & 31.5 & 30.2 & 29.5 & 29.2 \\ 
\midrule
\textbf{Vision-DeepResearch-30B} \\
Direct Answer 
&6.2 & 0.0 & 6.0 & 8.3 & 5.7 & 5.8 & 6.2 & 0.0 & 5.7 & 5.0 & 4.8 \\
CIS+TS &
35.4 & 43.1 & 44.0 & 33.3 & 32.1 & 32.7 & 35.4 & 36.8 & 35.8 & 52.5  & 37.8 \\

\bottomrule[1.2pt]
\end{tabular}
}
\end{table*}

\begin{table*}[ht]
\centering
\caption{
Performance comparison of models across different categories (Entity Recall) on testmini set.
CIS+TS means adopting both the cropped-image search and text search.
CIS+TS+MVF further applies the Multi-turn Visual Forcing strategy on top of CIS+TS.
}
\label{tab:accuracy-results-testmini}
\footnotesize
\setlength{\tabcolsep}{4pt}
\resizebox{0.95\linewidth}{!}{
\begin{tabular}{lccccccccccc}
\toprule[1.2pt]
Model / Setting & People & Object & Arch. & Nature & Sci\&Tech & Art\&Music & Sports & Movie & Game & Other & Overall \\
\midrule

\textbf{Gemini 2.5 Pro} \\
CIS+TS
& 10.4 & 11.8 & 12.0 & 10.4 & 11.3 & 11.5 & 14.6 & 14.0 & 11.3 & 25.0 & 13.0 \\
CIS+TS+MVF
& 27.1 & 39.2 & 32.0 & 41.7 & 28.3 & 40.4 & 22.9 & 28.1 & 24.5 & 37.5 & 32.0 \\

\midrule

\textbf{GPT-5} \\
CIS+TS
& 10.4 & 13.7 & 16.0 & 16.7 & 15.1 & 11.5 & 12.5 & 12.3 & 15.1 & 17.5 & 14.0 \\
CIS+TS+MVF
& 20.8 & 35.3 & 18.0 & 31.2 & 18.9 & 26.9 & 16.7 & 33.3 & 11.3 & 32.5 & 24.4 \\

\midrule

\textbf{Claude-4-Sonnet} \\
CIS+TS
& 6.2 & 7.8 & 6.0 & 6.2 & 7.5 & 7.7 & 6.2 & 8.8 & 7.5 & 17.5 & 8.0 \\
CIS+TS+MVF
& 25.0 & 17.6 & 22.0 & 25.0 & 18.9 & 19.2 & 14.6 & 12.3 & 11.3 & 30.0 & 19.2 \\

\midrule

\textbf{Qwen3-VL-30B-A3B-Instruct} \\
CIS+TS
& 10.4 & 11.8 & 14.0 & 12.5 & 13.2 & 13.5 & 10.4 & 12.3 & 15.1 & 17.5 & 13.0 \\
CIS+TS+MVF
& 14.6 & 25.5 & 16.0 & 22.9 & 17.0 & 21.2 & 12.5 & 26.3 & 3.8 & 45.0 & 20.0 \\

\midrule

\textbf{Qwen3-VL-235B-A22B-Instruct} \\
CIS+TS
& 14.6 & 9.8 & 16.0 & 33.3 & 13.2 & 19.2 & 16.7 & 14.0 & 22.6 & 17.5 & 17.6 \\
CIS+TS+MVF
& 31.2 & 27.5 & 32.0 & 27.1 & 26.4 & 25.0 & 16.7 & 36.8 & 24.5 & 37.5 & 28.4 \\

\midrule

\textbf{Vision-DeepResearch-8B} \\
CIS+TS
& 31.2 & 41.2 & 26.0 & 43.8 & 24.5 & 38.5 & 33.3 & 36.8 & 35.8 & 35.0 & 34.6 \\
\midrule

\textbf{Vision-DeepResearch-30B} \\
CIS+TS
& 41.7 & 51.0 & 52.0 & 39.6 & 39.6 & 40.4 & 41.7 & 43.9 & 43.4 & 60.0 & 45.0 \\

\bottomrule[1.2pt]
\end{tabular}
}
\end{table*}

\end{document}